\documentclass{article}

\usepackage[preprint]{neurips_2026}

\usepackage[utf8]{inputenc}
\usepackage[T1]{fontenc}
\usepackage[colorlinks=true,linkcolor=black,citecolor=black,urlcolor=blue]{hyperref}
\usepackage{url}
\usepackage{booktabs}
\usepackage{amsfonts}
\usepackage{amsmath}
\usepackage{nicefrac}
\usepackage{microtype}
\usepackage{xcolor}
\usepackage{graphicx}
\usepackage{subcaption}
\usepackage{multirow}
\usepackage{wrapfig}
\usepackage{algorithm}
\usepackage{algorithmic}
\usepackage{float}

\usepackage{listings}
\definecolor{codebg}{HTML}{F3EFFF}
\definecolor{codeframe}{HTML}{7C5CFC}
\definecolor{codecomment}{HTML}{6A6A6A}
\definecolor{codekw}{HTML}{7C5CFC}
\definecolor{codestr}{HTML}{C44E52}
\lstdefinestyle{purplebox}{
  backgroundcolor=\color{codebg},
  frame=single, rulecolor=\color{codeframe},
  basicstyle=\ttfamily\fontsize{7}{8.5}\selectfont,
  keywordstyle=\color{codekw}\bfseries,
  commentstyle=\color{codecomment}\itshape,
  stringstyle=\color{codestr},
  language=Python, showstringspaces=false,
  breaklines=true, columns=flexible, numbers=none,
  xleftmargin=2pt, xrightmargin=2pt,
  framexleftmargin=2pt, framexrightmargin=2pt,
  aboveskip=0.5em, belowskip=0.5em,
}

\title{Beyond Language: Format-Agnostic Reasoning Subspaces in Large Language Models}

\author{%
  Aojie Yuan$^{1}$ \quad Zhiyuan Su$^{2}$ \\[0.5em]
  $^1$University of Southern California \quad $^2$Duke University \\
  \texttt{aojieyua@usc.edu} \quad \texttt{zhiyuan.j.su@duke.edu}
}

\begin{document}

\maketitle

\begin{abstract}

Large language models represent the same reasoning in vastly different surface forms---English prose, Python code, mathematical notation---yet whether they share a common internal substrate across these symbolic systems remains unknown.
We introduce the TriForm Benchmark (18 concepts $\times$ 6 forms $\times$ 3 instances = 324 stimuli) and study five LLMs (1.6B--8B) across three architecture families.
Using permutation-corrected RSA, cross-form probing, and activation patching, we find converging evidence for a \emph{Format-Agnostic Reasoning Subspace} (FARS) in middle layers.
We make FARS concrete: concept-centroid PCA extracts a 10-dimensional subspace that amplifies concept structure $3\times$ while suppressing form information to near zero.
Replacing only these 10 dimensions during cross-form patching preserves 90--96\% of model output---far exceeding both full activation replacement (44--56\%) and variance-maximizing PCA (60--74\%)---while ablating them causes targeted disruption.
FARS generalizes to held-out concepts and, strikingly, converges across architectures (CCA $> 0.79$ for all model pairs), providing within-modality evidence for the Platonic Representation Hypothesis.
We further discover a \emph{declarative-procedural asymmetry}: representations are far more compatible between prose and mathematics than between either and code, suggesting that the critical axis of divergence is not linguistic vs.\ formal but declarative vs.\ procedural.

\end{abstract}

\section{Introduction}
\label{sec:intro}

The same logical inference---modus ponens, for instance---can be expressed as an English sentence, a Python function, or a line of mathematical notation. These surface forms share almost no tokens, obey different grammars, and employ different computational primitives, yet large language models process all of them. A growing body of evidence establishes that multilingual LLMs develop shared internal representations across natural languages~\citep{tang2024language, conneau2020emerging, dumas2025separating}, but natural languages share deep structural regularities---syntax, morphology, discourse connectives---making cross-lingual invariance a relatively mild test. Whether such invariance extends to \emph{genuinely different symbolic systems}---code, mathematical notation, structured data---remains an open question with implications for interpretability, safety, and our understanding of what these models learn about reasoning.

We investigate this question along three axes:

\begin{enumerate}
    \item \textbf{Existence.} Do LLMs develop representations invariant not just across \emph{languages} but across \emph{symbolic systems}---and can this shared structure be isolated as a concrete mathematical object?
    \item \textbf{Causality.} Is the shared structure causally involved in the model's computation, or merely a statistical regularity?
    \item \textbf{Universality.} Does the same subspace emerge across different architectures and training regimes, or is it an artifact of any particular model?
\end{enumerate}

These questions matter beyond scientific curiosity. If reasoning operates through a format-agnostic subspace, safety interventions discovered in one surface form should transfer to others; model capabilities are better understood as operating on an abstract computational substrate; and the Platonic Representation Hypothesis~\citep{huh2024platonic}---the conjecture that representations converge across modalities---finds support within a single modality.

Prior work has studied cross-\emph{lingual} invariance extensively. \citet{tang2024language} used activation entropy to show that language-specific neurons cluster in early and late layers, leaving middle layers relatively language-neutral. \citet{dumas2025separating} used activation patching to separate language from conceptual content at different layers. \citet{hu2025cross} demonstrated that knowledge-free reasoning transfers nearly perfectly across languages. These studies, however, only vary the \emph{language} while holding the symbolic system (natural language) constant. Whether the same invariance survives the jump to code and mathematics is the question our work addresses.

Using the \textbf{TriForm Benchmark}---324 controlled stimuli spanning 18 reasoning concepts in 6 surface forms across three invariance dimensions---and five LLMs (1.6B--8B) from three architecture families, we provide the following contributions:

\begin{enumerate}
    \item We present \textbf{converging correlational evidence} for a Format-Agnostic Reasoning Subspace (FARS) in middle layers, using permutation-corrected RSA, cross-form probing, and neuron-level entropy analysis (\S\ref{sec:correlational}).

    \item We establish \textbf{causal relevance} via cross-form activation patching and uncover a \emph{declarative-procedural asymmetry}: prose$\leftrightarrow$math patching preserves $4\times$ more structure than prose$\leftrightarrow$code (\S\ref{sec:causal}).

    \item We \textbf{extract FARS} as a concrete 10-dimensional subspace via concept-centroid PCA. Replacing only these 10 of 1600--4096 dimensions during cross-form patching preserves 90--96\% of model output, far exceeding both full replacement (44--56\%) and variance-maximizing PCA (60--74\%) (\S\ref{sec:extraction}).

    \item We demonstrate \textbf{universality}: FARS generalizes to held-out concepts and converges across architectures (CCA $> 0.79$), providing within-modality evidence for the Platonic Representation Hypothesis (\S\ref{sec:universality}).
\end{enumerate}

\section{Related Work}
\label{sec:related}

\paragraph{Cross-lingual representations in LLMs.}
A substantial literature documents that multilingual LLMs develop shared representations across languages. \citet{conneau2020emerging} showed that cross-lingual alignment emerges during pre-training of BLOOM. \citet{tang2024language} introduced LAPE (Language Activation Probability Entropy) to identify language-specific versus language-agnostic neurons, finding that language-specific neurons concentrate in the top and bottom layers while middle layers are more language-agnostic. \citet{ferrando2024similarity} demonstrated that syntactic circuits for subject-verb agreement in Gemma 2B are highly consistent across English and Spanish, with language-independent causal directions. \citet{liu2025middle} showed that middle-layer alignment objectives improve cross-lingual transfer across 1000+ language pairs. \citet{zhao2024multilingual} found that non-English representations become English-centric in middle layers. Our work extends this line of inquiry from the linguistic axis to the broader space of symbolic representations.

\paragraph{Activation patching and causal analysis.}
\citet{dumas2025separating} used activation patching on translation tasks to show that output language and conceptual content are encoded at different layers, with concepts representable in a language-agnostic form. \citet{anthropic2025biology} traced circuits in Claude 3.5 Haiku revealing language-independent internal reasoning with identifiable intermediate concepts. We adopt and extend activation patching to the cross-format setting, testing whether patching English activations into code or mathematical forward passes preserves reasoning.

\paragraph{Knowledge-free reasoning transfer.}
\citet{hu2025cross} decomposed LLM reasoning into knowledge retrieval and knowledge-free reasoning, showing that knowledge-free reasoning transfers nearly perfectly across languages with high hidden-state similarity and large neuron overlap. Our work tests whether this transfer extends beyond languages to symbolic systems.

\paragraph{Representation geometry and convergence.}
The Platonic Representation Hypothesis~\citep{huh2024platonic} posits that neural representations across modalities converge toward a shared statistical model of reality. \citet{park2024linear} formalized the linear representation hypothesis using counterfactuals, connecting it to linear probing and model steering. \citet{li2024world} showed that LLMs build goal-oriented state abstractions during planning tasks. Our work provides empirical evidence for convergence within a single modality across different symbolic encodings of the same reasoning content.

\paragraph{Format-agnostic reasoning.}
The closest work to ours is a concurrent study using RSA on the Cognitive Reflection Test (CRT) across scenario-based and mathematical formats~\citep{crosslingual2026rsa}, finding format-agnostic patterns in Qwen and Gemma. However, that study is limited to a single task type (CRT) and uses standard RSA without correcting for the non-independence of pairwise distances. \citet{chen2025abstract} identified language-agnostic shared neurons whose deactivation degrades performance across all languages, but restricted analysis to the cross-lingual axis. In the code domain, \citet{wan2025codellm} localized concept layers in code LLMs that encode language-agnostic representations across programming languages, finding smooth cross-programming-language transfer---an interesting contrast with our finding that natural-language-to-code transfer is much harder (declarative-procedural asymmetry). Our work is the first to systematically test three-dimensional format invariance (linguistic $\times$ symbolic $\times$ structural) with statistically rigorous methods, causal intervention, and concrete subspace extraction.

\section{Method}
\label{sec:method}

\begin{figure}[t]
\centering
\includegraphics[width=\linewidth]{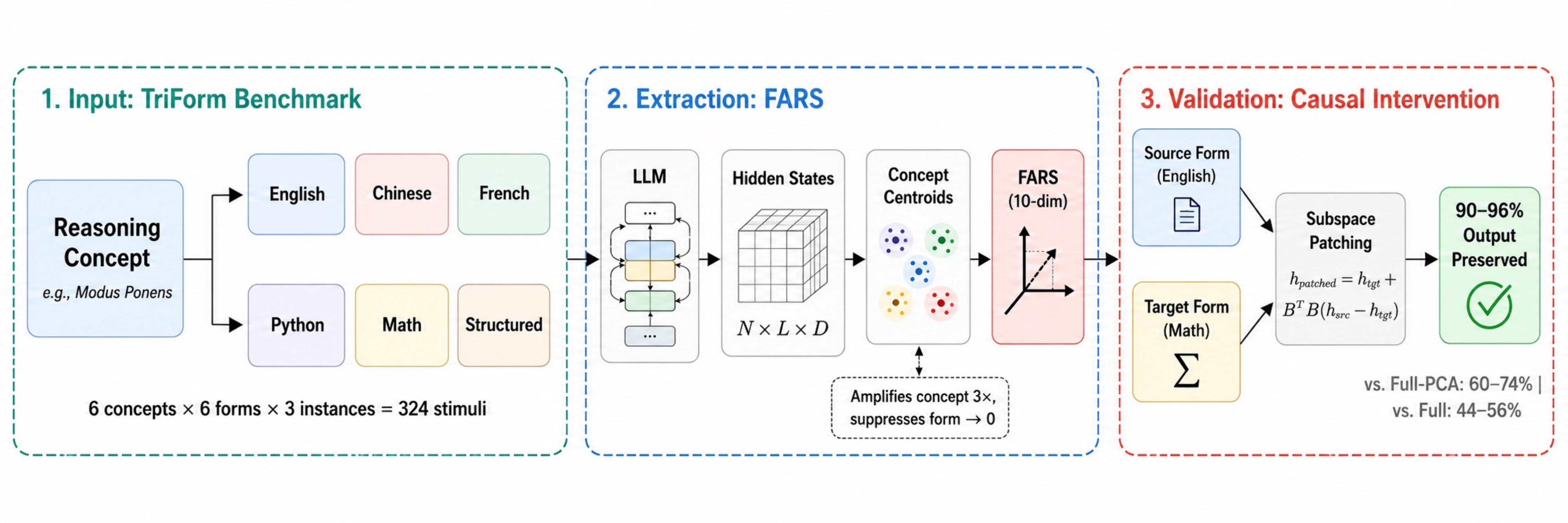}
\caption{\textbf{The FARS pipeline.} \textbf{Stage~1:} The TriForm Benchmark encodes 18 reasoning concepts in 6 surface forms. \textbf{Stage~2:} Concept-centroid PCA extracts a 10-dimensional FARS. \textbf{Stage~3:} Subspace patching replaces only FARS directions, preserving 90--96\% of output.}
\label{fig:pipeline}
\end{figure}

\subsection{TriForm Benchmark}
\label{sec:triform}

We construct a controlled benchmark of reasoning stimuli that vary systematically along three surface-form dimensions while holding semantic content constant.

\paragraph{Reasoning concepts.} We select 18 reasoning concepts spanning 5 domains: arithmetic (multi-step evaluation, modular arithmetic, proportional reasoning, GCD), logic (categorical syllogism, modus ponens, contrapositive, De Morgan's laws), relational (transitive ordering, set intersection, set difference, function composition), causal (causal chains, confounding variables, interventional reasoning), and spatial (directional composition, containment, mental rotation). Each concept represents a fundamental reasoning primitive that can be unambiguously expressed across symbolic systems.

\paragraph{Surface forms.} Each concept is expressed in 6 surface forms organized along three dimensions:
\begin{itemize}
    \item \textbf{Linguistic} (3 forms): English prose, Chinese prose, French prose---varying natural language while holding the symbolic system (natural language) constant.
    \item \textbf{Symbolic} (2 additional forms): Python code, mathematical notation---varying the symbolic system while holding the reasoning content constant.
    \item \textbf{Structural} (1 additional form): Structured step-by-step data---varying the organizational structure.
\end{itemize}

\paragraph{Instances.} Each concept has 3 semantically distinct instances (different numbers, entities, or configurations) to provide statistical power beyond a single exemplar. This yields $18 \times 3 \times 6 = 324$ total stimuli.

\paragraph{Semantic equivalence.} All surface forms for a given concept-instance pair are generated programmatically from a canonical specification, ensuring exact semantic equivalence. Each stimulus is self-contained (no context needed) and expresses both the reasoning structure and its conclusion.

\subsection{Representational Analysis}
\label{sec:rep_analysis}

\paragraph{Activation extraction.} For each stimulus, we extract the last-token hidden state at every transformer layer using forward hooks, yielding an activation tensor $\mathbf{A} \in \mathbb{R}^{N \times L \times D}$ where $N{=}324$ stimuli, $L$ is the number of layers, and $D$ is the hidden dimension. Last-token extraction is standard for decoder-only models, as this position has attended over the full input context.

\paragraph{Permutation-based RSA.}
Standard RSA computes Spearman correlation between empirical and theoretical representational dissimilarity matrices (RDMs), but treats the $\binom{N}{2}$ pairwise distances as independent observations---a severe violation when the same stimulus participates in $N{-}1$ pairs. We correct this by using permutation tests: we permute stimulus labels (rows and columns of the theoretical RDM simultaneously) 1{,}000 times, recomputing the Spearman correlation each time, and derive $p$-values from the empirical null distribution. We apply Benjamini-Hochberg FDR correction across all layers and RDM types.

We test four theoretical RDMs:
(a) \textbf{concept RDM}---same concept $=0$, different $=1$;
(b) \textbf{form RDM}---same surface form $=0$, different $=1$;
(c) \textbf{bias RDM}---continuous structural distances based on token count, entropy, and type-token ratio;
(d) \textbf{language-type RDM}---natural language $=0$, formal notation $=1$.

\paragraph{Ridge-regularized cross-form probing.}
For each source-target form pair, we train a ridge classifier ($\alpha{=}0.1$) on source-form activations to predict concept identity and evaluate on target-form activations. Ridge regularization addresses the extreme overfitting risk when fitting high-dimensional classifiers ($D{=}1600$--$4096$) on few samples (${\sim}54$ per form). We report the mean off-diagonal (cross-form) accuracy across all $\binom{6}{2}{=}15$ form pairs per layer.

\paragraph{Format-neuron entropy analysis.}
Inspired by LAPE~\citep{tang2024language}, we quantify how format-specific each neuron is. For each neuron $d$ at layer $\ell$, we compute the mean absolute activation per form, normalize to a distribution over forms, and compute the entropy:
\begin{equation}
H_{\ell,d} = -\sum_{f=1}^{F} p_f \log p_f, \quad \text{where } p_f = \frac{\bar{|a_f|}}{\sum_{f'} \bar{|a_{f'}|}}
\end{equation}
Neurons with high entropy ($H \approx \log F$) activate uniformly across forms (format-agnostic); those with low entropy activate preferentially for specific forms (format-specific). We report the fraction of neurons exceeding the 90th percentile entropy threshold per layer.

\paragraph{Dimension-wise CKA.}
To disentangle the three invariance dimensions, we compute linear Centered Kernel Alignment (CKA)~\citep{kornblith2019similarity} separately for linguistic pairs (EN$\leftrightarrow$ZH, EN$\leftrightarrow$FR, ZH$\leftrightarrow$FR), symbolic pairs (prose$\leftrightarrow$code, prose$\leftrightarrow$math, code$\leftrightarrow$math), and structural pairs (prose$\leftrightarrow$structured, code$\leftrightarrow$structured, math$\leftrightarrow$structured), averaging within each dimension.

\subsection{Subspace Extraction and Causal Intervention}
\label{sec:subspace_method}

To move FARS from hypothesis to concrete object, we extract it via concept-centroid PCA: at each layer, we average each concept's activations across all forms and instances to obtain 18 centroids in $\mathbb{R}^D$, then retain the top-$k$ principal components. As a control, form-centroid PCA (5 components) captures format-specific directions. Details in \S\ref{sec:extraction}.

We provide causal evidence through two interventions. \emph{Cross-form activation patching} replaces the last-token hidden state at layer $\ell$ with the activation from the same concept expressed in a different surface form, measuring top-10 token overlap between patched and clean outputs. \emph{Subspace patching} refines this by replacing only the FARS-projected component, testing whether the identified directions are causally sufficient. We test 4 form pairs across 50 concept instances at 8--9 layers per model.

\section{Experiments}
\label{sec:experiments}

\subsection{Setup}

\paragraph{Models.} We evaluate five decoder-only LLMs spanning an order of magnitude in parameter count: GPT-2 XL (1.6B, 48 layers), Qwen2.5-3B (3B, 36 layers), Qwen2.5-7B (7B, 28 layers), Mistral-7B-v0.3 (7B, 32 layers), and Llama-3.1-8B (8B, 32 layers). All models are evaluated in their base (non-instruction-tuned) form. This selection covers three model families (GPT, Qwen, Llama/Mistral) and enables both within-family and across-family comparisons.

\paragraph{Stimuli.} The TriForm Benchmark provides 324 stimuli: 18 concepts $\times$ 3 instances $\times$ 6 forms. Each form generates 54 stimuli distributed equally across all concepts.

\paragraph{Hardware.} Experiments are conducted on NVIDIA RTX 6000 Ada Generation GPUs (48GB VRAM). Activation extraction uses float16 precision; analysis uses float32.

\subsection{Correlational Evidence for FARS}
\label{sec:correlational}

We begin by asking whether concept-level structure exists in the representation space at all, using three complementary methods that probe different aspects of the geometry.

\paragraph{RSA reveals concept signal in middle layers.}
Table~\ref{tab:main_results} presents the key metrics across all five models. Concept RSA peaks in middle layers and strengthens with scale: $\rho_{\max} = 0.100 \to 0.200$. All values are significant after FDR correction ($p < 0.001$, permutation test). Form RSA dominates throughout ($\rho_{\text{form}} \approx 0.43$--$0.53$), confirming that surface form remains the primary organizing principle---but concept signal grows faster with scale. Notably, Mistral-7B achieves the highest concept RSA ($\rho{=}0.200$), suggesting that training data composition matters beyond parameter count.

\begin{table}[t]
\centering
\caption{Summary of representational analysis. RSA-C: concept RSA peak; RSA-F: form RSA peak; Probe: cross-form probing accuracy (chance = 5.6\%); Agn\%: peak format-agnostic neuron fraction. All RSA $p < 0.001$ (permutation test, FDR-corrected).}
\label{tab:main_results}
\vspace{0.5em}
\begin{tabular}{@{}lccccccc@{}}
\toprule
\textbf{Model} & \textbf{Params} & \textbf{Layers} & \textbf{RSA-C $\rho$ (layer)} & \textbf{RSA-F $\rho$} & \textbf{Probe\%} & \textbf{Agn\% (layer)} \\
\midrule
GPT-2 XL & 1.6B & 48 & 0.100 (19) & 0.430 & 32.7 & 20.0 (23) \\
Qwen2.5-3B & 3B & 36 & 0.126 (21) & 0.501 & 52.8 & 23.3 (22) \\
Qwen2.5-7B & 7B & 28 & 0.150 (13) & 0.485 & 59.8 & 19.5 (15) \\
Mistral-7B & 7B & 32 & 0.200 (8) & 0.529 & 60.3 & 20.8 (11) \\
Llama-3.1-8B & 8B & 32 & 0.165 (9) & 0.525 & 53.9 & 21.1 (10) \\
\bottomrule
\end{tabular}
\end{table}

\paragraph{Cross-form probing confirms concept identity survives form transformation.}
Cross-form probing accuracy far exceeds chance (5.6\% for 18-way classification) in all models, reaching 60.3\% for Mistral-7B. The jump from GPT-2 XL (32.7\%) to Qwen2.5-3B (52.8\%) suggests a qualitative shift in format-agnostic encoding between 1.6B and 3B parameters. Among 7B models, Mistral-7B substantially outperforms the larger Llama-3.1-8B (53.9\%), highlighting the role of training data diversity.

\begin{figure}[t]
\centering
\includegraphics[width=\linewidth]{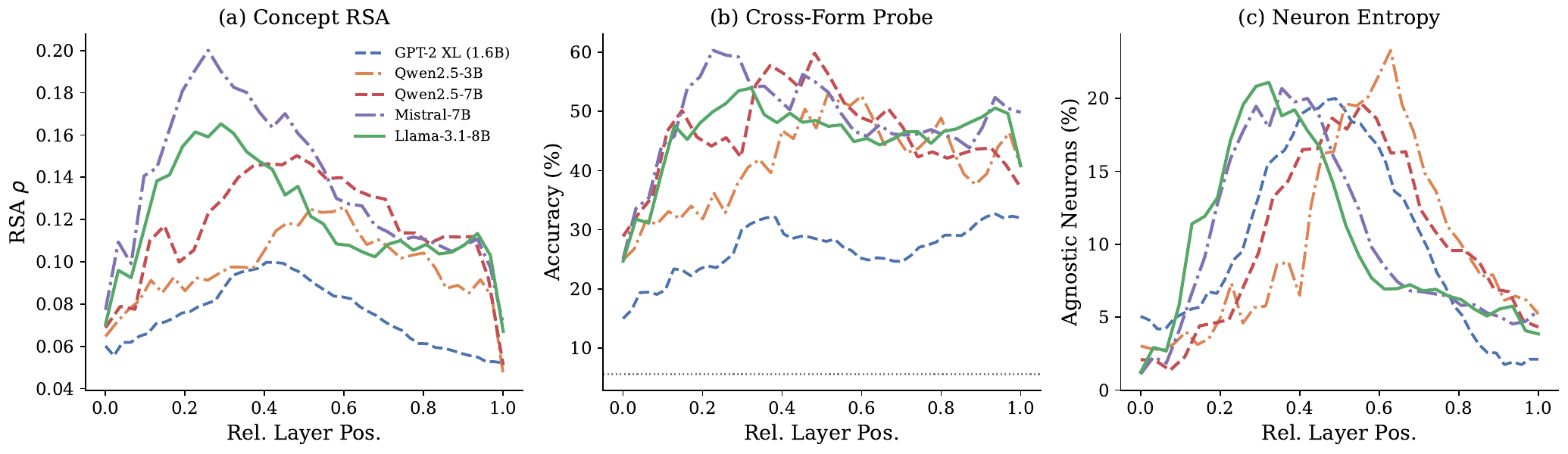}
\caption{Three complementary views of format-agnostic representations across layers. (a)~Concept RSA peaks in middle layers and strengthens with model scale. (b)~Cross-form probing accuracy far exceeds chance (5.6\%). (c)~Format-agnostic neuron fraction concentrates in middle layers, mirroring the language-specific neuron distribution of \citet{tang2024language}.}
\label{fig:main_results}
\end{figure}

\paragraph{Format-agnostic neurons concentrate in middle layers.}
Figure~\ref{fig:main_results} presents the three findings together. All five models exhibit the same spatial pattern: a unimodal peak of format-agnostic neurons in middle layers (20--23\% of neurons) with format-specific neurons dominating early (embedding/tokenization) and late (generation) layers. This mirrors the language-specific neuron distribution found by \citet{tang2024language} but extends it to the full format axis including code and mathematical notation.

\paragraph{Invariance converges across dimensions at scale.}
Dimension-wise CKA (Table~\ref{tab:dimension_cka}) reveals that linguistic, symbolic, and structural invariance all improve with scale and \emph{converge}: Llama-3.1-8B shows the tightest spread ($0.018$). Strikingly, symbolic invariance (prose vs.\ code vs.\ math) approaches linguistic invariance in larger models, despite fundamental differences in tokenization and syntax (Appendix Figure~\ref{fig:cka_dimensions}).

\begin{table}[t]
\centering
\caption{Peak CKA alignment by invariance dimension. All three dimensions improve with scale and converge.}
\label{tab:dimension_cka}
\vspace{0.5em}
\begin{tabular}{@{}lcccc@{}}
\toprule
\textbf{Model} & \textbf{Linguistic} & \textbf{Symbolic} & \textbf{Structural} & \textbf{Spread} \\
\midrule
GPT-2 XL (1.6B) & 0.751 & 0.756 & 0.726 & 0.030 \\
Qwen2.5-3B & 0.812 & 0.775 & 0.768 & 0.044 \\
Qwen2.5-7B & 0.844 & 0.817 & 0.792 & 0.052 \\
Mistral-7B & 0.867 & 0.833 & 0.823 & 0.043 \\
Llama-3.1-8B & 0.835 & 0.819 & 0.817 & 0.018 \\
\bottomrule
\end{tabular}
\end{table}

These three analyses converge: concept-level structure exists in middle layers, is linearly decodable across form boundaries, and is carried by a spatially localized subset of neurons. But correlation does not imply causation. We next ask whether this structure is causally relevant.

\subsection{Causal Evidence: Activation Patching}
\label{sec:causal}

\begin{figure}[t]
\centering
\includegraphics[width=\linewidth]{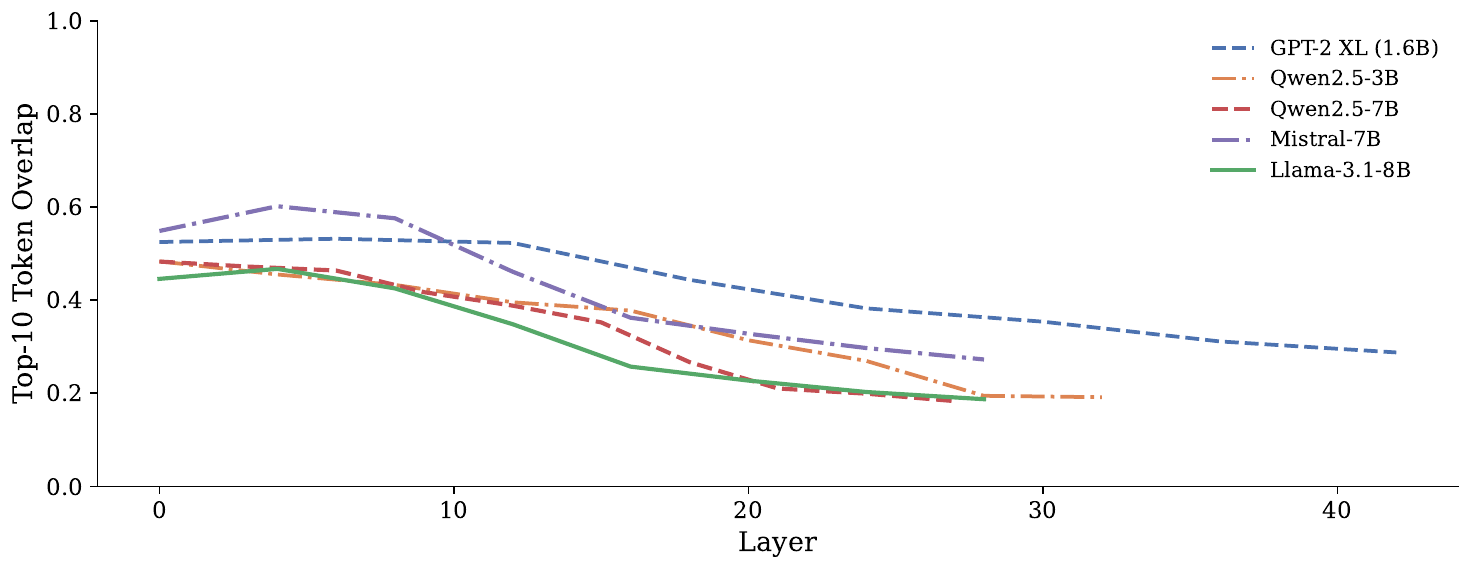}
\caption{Cross-form patching overlap by layer. Patching is most effective at early-to-middle layers and degrades toward later layers where form-specific generation dominates.}
\label{fig:patching_layers}
\end{figure}

\paragraph{Cross-form patching.}
We replace last-token activations from a source form (e.g., English prose) into the forward pass of a target form (e.g., mathematical notation) at each layer and measure top-10 token overlap with the clean target output. Table~\ref{tab:patching} summarizes the results. Patching EN$\to$Math consistently preserves 65--73\% of top-token predictions, indicating strong representational compatibility between declarative forms. Patching degrades monotonically from early to late layers (Figure~\ref{fig:patching_layers}), consistent with the correlational finding that middle layers encode format-agnostic content.

\begin{table}[t]
\centering
\caption{Activation patching: mean top-10 token overlap, averaged across layers. EN$\to$Math compatibility is $3$--$4\times$ higher than EN$\to$Code, revealing a declarative-procedural asymmetry.}
\label{tab:patching}
\vspace{0.5em}
\begin{tabular}{@{}lcccc@{}}
\toprule
\textbf{Model} & \textbf{EN$\to$Math} & \textbf{EN$\to$ZH} & \textbf{EN$\to$Code} & \textbf{Code$\to$Math} \\
\midrule
GPT-2 XL (1.6B) & 0.708 & 0.496 & 0.221 & 0.255 \\
Qwen2.5-3B & 0.684 & 0.343 & 0.216 & 0.141 \\
Qwen2.5-7B & 0.654 & 0.376 & 0.186 & 0.158 \\
Mistral-7B & 0.726 & 0.437 & 0.215 & 0.345 \\
Llama-3.1-8B & 0.677 & 0.272 & 0.162 & 0.170 \\
\bottomrule
\end{tabular}
\end{table}

\paragraph{Declarative-procedural asymmetry.}
The most striking pattern in Table~\ref{tab:patching} is the gap between EN$\to$Math (0.65--0.73) and EN$\to$Code (0.16--0.22)---a 3--4$\times$ difference that is remarkably stable across all five models. English prose and mathematical notation share substantial representational structure despite using entirely different symbolic systems, likely because both describe reasoning \emph{declaratively}. Code encodes the same reasoning \emph{procedurally} (control flow, variable assignment), creating a fundamentally larger representational gap. This suggests that the critical axis of divergence in LLM representations is not linguistic vs.\ formal but \emph{declarative vs.\ procedural}---a distinction not previously identified in the cross-lingual literature (see Appendix Figure~\ref{fig:patching_formpairs} for layer-wise breakdown).

\begin{figure}[t]
\centering
\includegraphics[width=\linewidth]{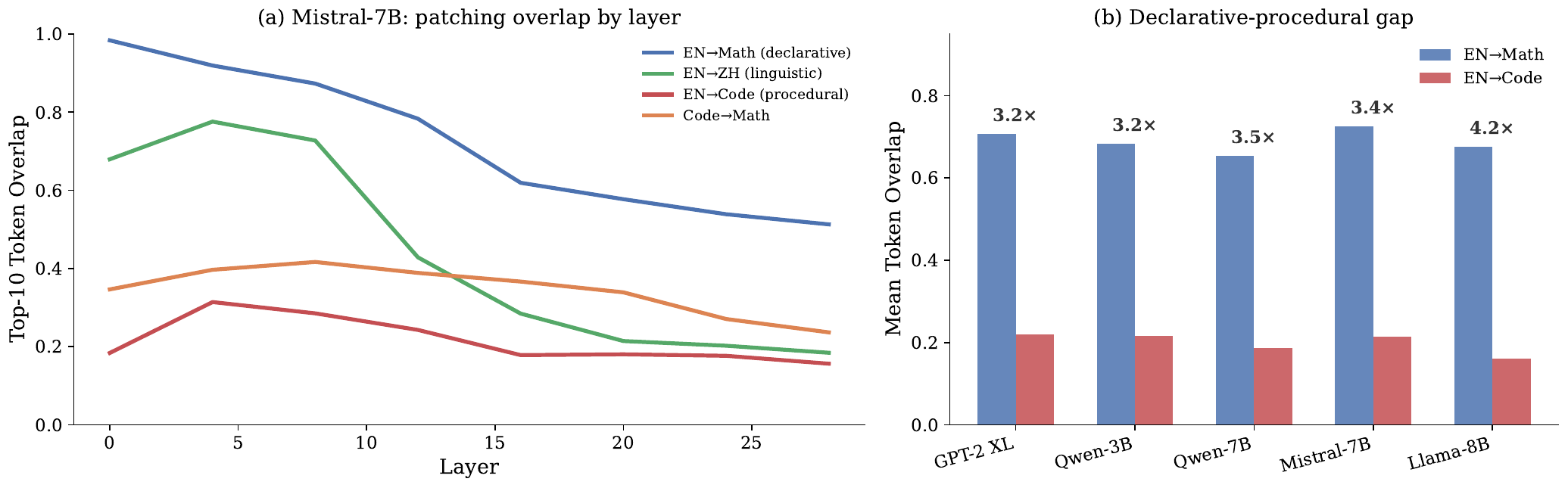}
\caption{Declarative-procedural asymmetry. \textbf{(a)}~Layer-wise patching overlap for Mistral-7B. \textbf{(b)}~The EN$\to$Math / EN$\to$Code ratio is $3$--$4\times$ across all five models.}
\label{fig:decl_proc}
\end{figure}

Patching establishes causal relevance of the shared representation, but the intervention is coarse: it replaces \emph{all} dimensions at a given layer. Can we identify and isolate the specific directions that carry concept information across forms?

\subsection{Extracting and Validating FARS}
\label{sec:extraction}

\paragraph{From hypothesis to object.}
We extract FARS via concept-centroid PCA: at each layer, we average each concept's activations across all 6 forms and 3 instances to obtain 18 centroids in $\mathbb{R}^D$, then take the top-$k$ principal components. This recovers directions along which concepts differ while form-specific variation is averaged out. As a control, form-centroid PCA (5 components) captures format-\emph{specific} information.

\begin{figure}[t]
\centering
\includegraphics[width=\linewidth]{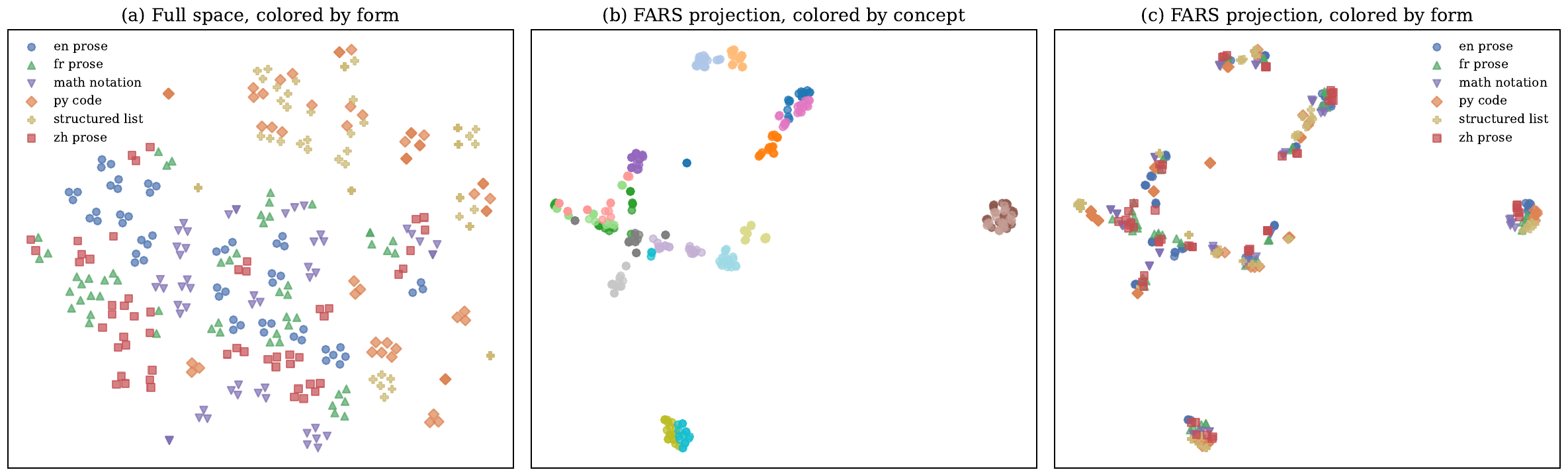}
\caption{t-SNE visualization (Mistral-7B, best FARS layer). \textbf{(a)}~Full space: stimuli cluster by form. \textbf{(b)}~FARS projection by concept: 18 clusters emerge. \textbf{(c)}~FARS by form: forms intermixed---FARS has factored concept from form.}
\label{fig:tsne}
\end{figure}

Table~\ref{tab:subspace} confirms the extraction succeeds: concept RSA increases 2--3$\times$ (e.g., $0.100 \to 0.317$ for GPT-2 XL), while form RSA drops to near zero. Cross-form probe accuracy improves by up to 20 percentage points. The control subspace shows the exact opposite: high form RSA (${\sim}0.63$) with concept RSA at zero, confirming complementary information capture. Figure~\ref{fig:tsne} provides a striking visualization: in the full space stimuli cluster by form; after FARS projection they cluster by concept with forms intermixed.

\begin{table}[t]
\centering
\caption{Subspace extraction results. FARS (concept-centroid PCA, 10 dims) amplifies concept signal and suppresses form signal; the control (form-centroid PCA, 5 dims) shows the opposite.}
\label{tab:subspace}
\vspace{0.5em}
\begin{tabular}{@{}lcccc|cc@{}}
\toprule
 & \multicolumn{4}{c|}{\textbf{FARS Subspace}} & \multicolumn{2}{c}{\textbf{Control}} \\
\textbf{Model} & RSA-C & RSA-F & Probe\% & Boost & RSA-F & RSA-C \\
\midrule
GPT-2 XL (1.6B) & 0.317 & 0.118 & 52.8 & $+$20.1 & 0.626 & $-$0.003 \\
Qwen2.5-3B & 0.357 & 0.056 & 64.4 & $+$11.6 & 0.634 & $-$0.005 \\
Qwen2.5-7B & 0.367 & 0.009 & 68.0 & $+$8.2 & 0.638 & $-$0.010 \\
Mistral-7B & 0.366 & 0.008 & 70.5 & $+$10.2 & 0.631 & $-$0.003 \\
Llama-3.1-8B & 0.368 & 0.015 & 71.9 & $+$18.0 & 0.634 & $-$0.007 \\
\bottomrule
\end{tabular}
\end{table}

\paragraph{Causal validation via subspace patching.}
Extraction alone is correlational. We provide causal evidence with two targeted interventions. \emph{Subspace patching} replaces only the FARS-projected component during cross-form patching:
\begin{equation}
\mathbf{h}_{\text{patched}} = \mathbf{h}_{\text{tgt}} + \mathbf{B}^\top \mathbf{B}\, (\mathbf{h}_{\text{src}} - \mathbf{h}_{\text{tgt}})
\label{eq:subspace_patch}
\end{equation}
where $\mathbf{B} \in \mathbb{R}^{k \times D}$ are the FARS basis vectors. \emph{Subspace ablation} removes FARS directions:
\begin{equation}
\mathbf{h}_{\text{abl}} = (\mathbf{I} - \mathbf{B}^\top \mathbf{B})\, \mathbf{h}
\end{equation}

Table~\ref{tab:subspace_patching} presents the results alongside three controls. The interpretation requires understanding a \emph{disruption-compatibility trade-off}: replacing 10 dimensions has two effects---it disrupts whatever information those dimensions carried, and it introduces source-form information that may or may not be compatible with the target's downstream computation. This trade-off explains the ordering:

\begin{itemize}
    \item \textbf{Random (${{\sim}}99\%$):} Random directions carry negligible structured information; replacing them neither disrupts nor transfers anything meaningful. This is a near-no-op, confirming that 10 of 1600--4096 dimensions is a small intervention.
    \item \textbf{FARS ($90$--$96\%$):} These directions carry concept information (ablation KL: $0.05$--$0.43$), so the intervention is \emph{not} a no-op. Yet overlap remains very high because the concept information being transplanted is \emph{cross-form compatible}---precisely what FARS predicts.
    \item \textbf{Full-PCA ($60$--$74\%$):} Top variance directions are dominated by form-specific features. Replacing them introduces form-\emph{incompatible} information, causing substantial disruption.
    \item \textbf{Full replacement ($44$--$56\%$):} Replacing all dimensions maximizes disruption from form-incompatible directions.
\end{itemize}

\begin{figure}[t]
\centering
\includegraphics[width=\linewidth]{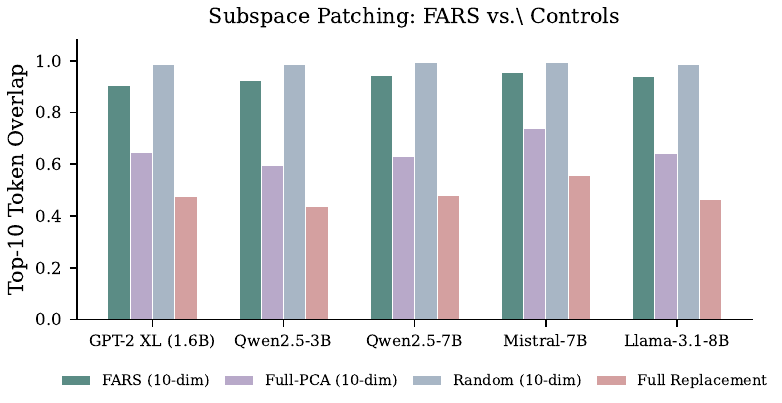}
\caption{Subspace patching: four-way comparison. FARS directions carry causal concept information yet remain cross-form compatible, while Full-PCA directions carry form-specific information that is incompatible across forms. Random 10-dim is a near-no-op baseline.}
\label{fig:subspace_comparison}
\end{figure}

The critical comparison is FARS vs.\ Full-PCA: both replace exactly 10 dimensions, but FARS selects concept-compatible directions while Full-PCA selects form-heavy (and thus cross-form-incompatible) directions, resulting in a 30-percentage-point gap. Figure~\ref{fig:subspace_comparison} visualizes this four-way comparison.

\begin{table}[t]
\centering
\caption{Causal validation of FARS. Patching overlap ($\uparrow$): fraction of top-10 tokens preserved. Ablation KL ($\downarrow$): divergence from removing subspace directions. 95\% CIs via concept-block bootstrap (5{,}000 resamples).}
\label{tab:subspace_patching}
\vspace{0.5em}
\begin{tabular}{@{}lcccc|cc@{}}
\toprule
 & \multicolumn{4}{c|}{\textbf{Patching Overlap} $\uparrow$} & \multicolumn{2}{c}{\textbf{Ablation KL} $\downarrow$} \\
\textbf{Model} & FARS & PCA-10 & Full & 95\% CI (FARS) & FARS & Form \\
\midrule
GPT-2 XL (1.6B) & \textbf{0.904} & 0.645 & 0.474 & [0.82, 0.89] & 0.430 & 0.704 \\
Qwen2.5-3B & \textbf{0.922} & 0.595 & 0.436 & [0.87, 0.90] & 0.125 & 1.951 \\
Qwen2.5-7B & \textbf{0.943} & 0.632 & 0.478 & [0.87, 0.90] & 0.144 & 2.017 \\
Mistral-7B & \textbf{0.955} & 0.740 & 0.557 & [0.91, 0.94] & 0.054 & 0.886 \\
Llama-3.1-8B & \textbf{0.941} & 0.641 & 0.465 & [0.89, 0.92] & 0.063 & 1.315 \\
\bottomrule
\end{tabular}
\end{table}

\begin{figure}[t]
\centering
\includegraphics[width=\linewidth]{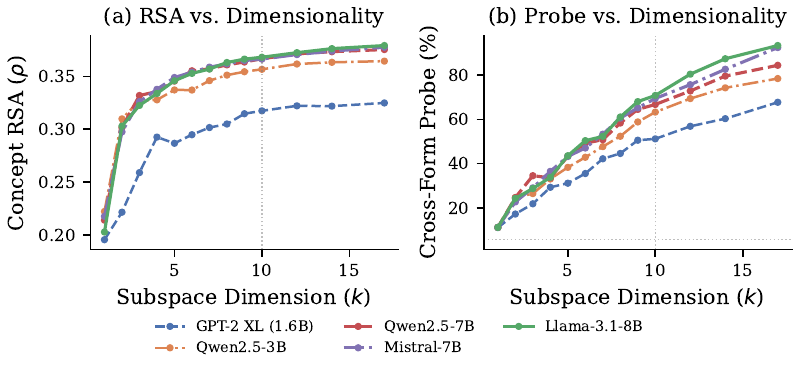}
\caption{Dimensionality sweep. Concept RSA plateaus around $k{=}8$, confirming FARS is genuinely low-dimensional. Vertical gray line marks $k{=}10$.}
\label{fig:dim_sweep}
\end{figure}

\paragraph{Intrinsic dimensionality.}
We vary $k$ from 1 to 17. Figure~\ref{fig:dim_sweep} shows that concept RSA rises steeply to $k{=}5$ and plateaus by $k{=}8$--$10$, with the first 10 components capturing 85--87\% of inter-concept variance. FARS is genuinely low-dimensional: the reasoning-relevant structure lives in a ${\sim}5$--$10$ dimensional manifold within a $1600$--$4096$ dimensional space.

\subsection{Universality of FARS}
\label{sec:universality}

A subspace extracted from a specific set of concepts on a specific model could be an artifact of overfitting. We test universality along two axes: concept generalization and cross-model convergence.

\begin{figure}[t]
\centering
\includegraphics[width=\linewidth]{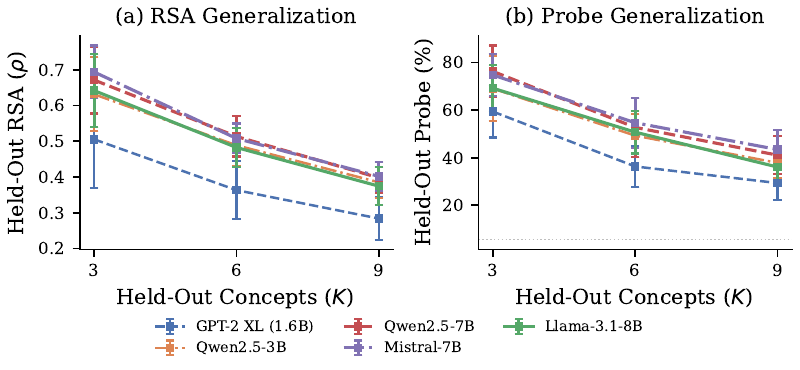}
\caption{Generalization: FARS extracted from $18{-}K$ concepts, evaluated on $K$ held-out concepts. Performance degrades gracefully.}
\label{fig:generalization}
\end{figure}

\paragraph{Generalization to held-out concepts.}
We hold out $K \in \{3, 6, 9\}$ concepts, extract FARS from the remaining $18{-}K$, and evaluate on the held-out projections. Figure~\ref{fig:generalization} shows graceful degradation: even at $K{=}9$ (50\% held out), held-out concept RSA remains $0.28$--$0.40$ and probe accuracy reaches $29$--$43\%$ (chance $= 5.6\%$)---well above chance and comparable to full-space RSA ($0.10$--$0.20$). FARS captures general reasoning structure, not concept-specific patterns.

\begin{figure}[t]
\centering
\includegraphics[width=0.55\linewidth]{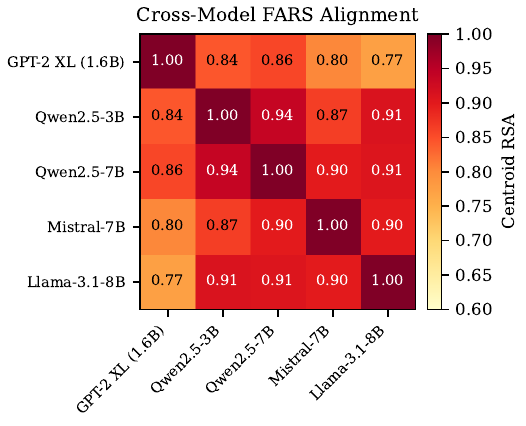}
\caption{Cross-model FARS alignment (centroid RSA). Same-family pairs exceed $0.90$; cross-family pairs exceed $0.77$, suggesting convergent reasoning geometry.}
\label{fig:alignment}
\end{figure}

\paragraph{Cross-model convergence.}
Do different architectures converge on the same FARS? We project all 324 stimuli onto each model's FARS (at its best layer) and compute pairwise CCA and centroid RSA. Figure~\ref{fig:alignment} reveals striking convergence: same-family models (Qwen-3B vs.\ Qwen-7B) achieve CCA $> 0.90$ and centroid RSA $> 0.93$, while cross-family pairs (GPT-2 vs.\ Llama) still reach CCA $> 0.79$ and centroid RSA $> 0.77$---despite different architectures, training corpora, and tokenizers. FARS is not an artifact of any particular model but a \emph{universal} geometric structure, providing within-modality evidence for the Platonic Representation Hypothesis~\citep{huh2024platonic}.

\section{Discussion}
\label{sec:discussion}

\paragraph{The form-dominance paradox.}
A natural objection to format-agnostic representations is that global RDMs are overwhelmingly form-dominated ($\rho_{\text{form}} \gg \rho_{\text{concept}}$). Our results resolve this apparent contradiction: FARS occupies only 10 of 1600--4096 dimensions. Surface form controls the vast majority of the representational space; concept identity is confined to a low-rank component that is invisible to global similarity measures but recoverable via concept-centroid PCA. The concept signal is real---it is simply embedded in a much larger form-structured space.

\paragraph{The declarative-procedural boundary.}
The $3$--$4\times$ gap between EN$\to$Math and EN$\to$Code patching refines the format-invariance hypothesis in an unexpected direction. LLMs develop strong invariance between symbolic systems that describe reasoning \emph{declaratively} (prose, mathematics, structured data) but substantially weaker invariance with systems that encode reasoning \emph{procedurally} (code). This aligns with an instructive contrast in the literature: \citet{wan2025codellm} found smooth transfer \emph{between} programming languages, confirming that procedural representations are internally coherent. The gap lies specifically at the declarative-procedural boundary---not between individual formats, but between computational paradigms.

\paragraph{Implications for representation engineering.}
FARS provides a concrete, cheaply extractable target for representation-based interventions. The 10-dimensional basis is obtained via PCA on 18 centroids---a computation that takes seconds. Steering vectors computed in one surface form should transfer along FARS directions to other declarative forms; sparse autoencoder features within FARS may be particularly interpretable as concept-level abstractions, while features in the orthogonal complement should encode tokenization and syntax. More practically, monitoring FARS alignment during training could predict cross-format generalization before behavioral evaluation, consistent with the linear representation hypothesis formalized by \citet{park2024linear}.

\paragraph{Safety considerations.}
The partial nature of format invariance has direct safety implications. Most alignment and red-teaming work operates in English prose---a declarative format. Our results indicate that such interventions should transfer to mathematical notation but may be substantially less effective for code. This asymmetry represents a potential attack surface: harmful reasoning reformulated as code may bypass prose-trained safety filters. FARS extraction offers a mechanistic diagnostic for anticipating such transfer failures.

\paragraph{Limitations.}
Several limitations qualify our conclusions.
(1)~We evaluate base models only; instruction tuning and RLHF may reshape FARS geometry in ways we have not characterized.
(2)~FARS extraction requires concept labels and is therefore supervised; unsupervised discovery of format-agnostic directions (e.g., via contrastive learning or dictionary methods) remains an open challenge.
(3)~The TriForm Benchmark uses programmatically generated stimuli; naturally occurring reasoning may exhibit more complex format interactions.
(4)~Our largest model is 8B parameters; whether FARS dimensionality scales with model size at 70B+ is unknown.
(5)~The structural invariance dimension has only one form (structured data), making the three-way convergence claim weakest along this axis.
(6)~We characterize FARS at the subspace level but have not traced the underlying circuits (attention heads, MLP neurons) that implement it.

\section{Conclusion}
\label{sec:conclusion}

We investigated whether LLMs develop representations invariant across symbolic systems---not merely across natural languages. Across five models from three architecture families, we find affirmative evidence along all three axes: a Format-Agnostic Reasoning Subspace exists in middle layers, is causally sufficient for cross-form reasoning (90--96\% output preservation from 10 dimensions), and converges across independently trained architectures (CCA $> 0.79$).

Our results also reveal a principled boundary to this invariance. The declarative-procedural asymmetry---a $3$--$4\times$ gap between prose$\leftrightarrow$math and prose$\leftrightarrow$code transfer, stable across all models---indicates that LLMs maintain distinct representational regimes for declarative and procedural encodings of the same reasoning content. This boundary, to our knowledge not previously characterized, has implications for cross-format safety and for understanding the computational structure of learned representations.

FARS provides not only a theoretical contribution but a practical tool: a set of basis vectors, extractable via PCA on concept centroids, that isolates reasoning content from surface form at each layer. We anticipate that this object will serve as a foundation for representation engineering, cross-format safety auditing, and circuit-level analysis of abstract reasoning in LLMs.




\bibliographystyle{plainnat}

\begin{thebibliography}{20}

\bibitem[Chen et al.(2025)]{chen2025abstract}
Yuxin Chen, Yiran Zhao, Yang Zhang, An Zhang, Kenji Kawaguchi, Shafiq Joty, Junnan Li, Tat-Seng Chua, Michael Qizhe Shieh, and Wenxuan Zhang.
\newblock The emergence of abstract thought in large language models beyond any language.
\newblock In \emph{NeurIPS}, 2025.

\bibitem[Wu et al.(2020)]{conneau2020emerging}
Shijie Wu, Alexis Conneau, Haoran Li, Luke Zettlemoyer, and Veselin Stoyanov.
\newblock Emerging cross-lingual structure in pretrained language models.
\newblock In \emph{ACL}, 2020.

\bibitem[Dumas et al.(2025)]{dumas2025separating}
Clement Dumas, Chris Wendler, Veniamin Veselovsky, Giovanni Monea, and Robert West.
\newblock Separating tongue from thought: Activation patching reveals language-agnostic concept representations in transformers.
\newblock In \emph{ACL}, 2025.

\bibitem[Ferrando \& Costa-juss{\`a}(2024)]{ferrando2024similarity}
Javier Ferrando and Marta R.~Costa-juss{\`a}.
\newblock On the similarity of circuits across languages: A case study on the subject-verb agreement task.
\newblock In \emph{Findings of EMNLP}, 2024.

\bibitem[Hu et al.(2025)]{hu2025cross}
Peng Hu, Sizhe Liu, Changjiang Gao, Xin Huang, Xue Han, Junlan Feng, Chao Deng, and Shujian Huang.
\newblock Large language models are cross-lingual knowledge-free reasoners.
\newblock In \emph{NAACL}, 2025.

\bibitem[Huh et al.(2024)]{huh2024platonic}
Minyoung Huh, Brian Cheung, Tongzhou Wang, and Phillip Isola.
\newblock Position: The Platonic representation hypothesis.
\newblock In \emph{ICML}, 2024.

\bibitem[Kornblith et al.(2019)]{kornblith2019similarity}
Simon Kornblith, Mohammad Norouzi, Honglak Lee, and Geoffrey Hinton.
\newblock Similarity of neural network representations revisited.
\newblock In \emph{ICML}, 2019.

\bibitem[Li et al.(2024)]{li2024world}
Zichao Li, Yanshuai Cao, and Jackie CK~Cheung.
\newblock Do {LLMs} build world representations? {P}robing through the lens of state abstraction.
\newblock In \emph{NeurIPS}, 2024.

\bibitem[Lindsey et al.(2025)]{anthropic2025biology}
Jack Lindsey et al.
\newblock On the biology of a large language model.
\newblock \emph{Transformer Circuits}, 2025.

\bibitem[Liu \& Niehues(2025)]{liu2025middle}
Danni Liu and Jan Niehues.
\newblock Middle-layer representation alignment for cross-lingual transfer in fine-tuned {LLMs}.
\newblock In \emph{ACL}, 2025.

\bibitem[Park et al.(2024)]{park2024linear}
Kiho Park, Yo Joong Choe, and Victor Veitch.
\newblock The linear representation hypothesis and the geometry of large language models.
\newblock In \emph{ICML}, 2024.

\bibitem[Tang et al.(2024)]{tang2024language}
Tianyi Tang, Wenyang Luo, Haoyang Huang, Dongdong Zhang, Xiaolei Wang, Xin Zhao, Furu Wei, and Ji-Rong Wen.
\newblock Language-specific neurons: The key to multilingual capabilities in large language models.
\newblock In \emph{ACL}, 2024.

\bibitem[Wang et al.(2024)]{crosslingual2026rsa}
Zining Wang, Jiaqi Li, and Yan Cong.
\newblock The reasoning-like capabilities of large language models across different languages: Insights from representational similarity analysis.
\newblock \emph{Computers in Human Behavior: AI}, 2024.

\bibitem[Zhao et al.(2024)]{zhao2024multilingual}
Yiran Zhao, Wenxuan Zhang, Guizhen Chen, Kenji Kawaguchi, and Lidong Bing.
\newblock How do large language models handle multilingualism?
\newblock In \emph{NeurIPS}, 2024.

\bibitem[Yin et al.(2025)]{wan2025codellm}
Zhe Yin, Xiaodong Gu, and Beijun Shen.
\newblock Neuron-guided interpretation of code LLMs: Where, why, and how?
\newblock In \emph{FSE}, 2026. arXiv:2512.19980.

\end{thebibliography}

\appendix

\section{Stimulus Examples}
\label{app:stimuli}

Table~\ref{tab:stimulus_example} shows one reasoning concept (modus ponens) expressed across all 6 surface forms.

\begin{table}[H]
\centering
\caption{Example stimulus: modus ponens in 6 surface forms.}
\label{tab:stimulus_example}
\small
\begin{tabular}{@{}lp{10cm}@{}}
\toprule
\textbf{Form} & \textbf{Text} \\
\midrule
English & If it rains then the ground is wet. It rains. Therefore the ground is wet. \\
Chinese & \emph{[Equivalent reasoning in Mandarin Chinese]} \\
French & S'il pleut alors le sol est mouill\'{e}. Il pleut. Donc le sol est mouill\'{e}. \\
Python & \texttt{def modus\_ponens(p, q): return q if p else None} \\
Math & $P \to Q,\; P \;\vdash\; Q$ \\
Structured & \texttt{P1: P->Q | P2: P | Rule: MP | Conclusion: Q} \\
\bottomrule
\end{tabular}
\end{table}

\section{TriForm Benchmark Concept Inventory}
\label{app:benchmark}

\begin{table}[H]
\centering
\caption{Complete concept inventory: 18 concepts across 5 reasoning domains.}
\small
\begin{tabular}{@{}llp{6cm}@{}}
\toprule
\textbf{Domain} & \textbf{Concept} & \textbf{Example (English)} \\
\midrule
Arithmetic & Multi-step evaluation & Calculate $(7{+}3){\times}4{-}8/2$ \\
 & Modular arithmetic & What is $47 \bmod 7$? \\
 & Proportional reasoning & 3 items cost \$12; how much for 7? \\
 & GCD (Euclidean) & Find $\gcd(48, 18)$ \\
\midrule
Logic & Categorical syllogism & All A are B; all C are A; therefore\ldots \\
 & Modus ponens & If P then Q; P; therefore Q \\
 & Contrapositive & If P then Q; not Q; therefore not P \\
 & De Morgan's laws & $\neg(A \wedge B) = (\neg A) \vee (\neg B)$ \\
\midrule
Relational & Transitive ordering & A $>$ B and B $>$ C; who is greatest? \\
 & Set intersection & $A \cap B$ \\
 & Set difference & $A \setminus B$ \\
 & Function composition & $f(g(x))$ \\
\midrule
Causal & Causal chain & A$\to$B$\to$C; does A cause C? \\
 & Confounding & A$\leftarrow$C$\to$B; does A cause B? \\
 & Interventional & $P(Y|\text{do}(X))$ vs.\ $P(Y|X)$ \\
\midrule
Spatial & Direction composition & A north of B, B east of C; A from C? \\
 & Containment & A inside B, B inside C; A inside C? \\
 & Mental rotation & Rotate 90\textdegree; where is the vertex? \\
\bottomrule
\end{tabular}
\end{table}

\section{Subspace Extraction Details}
\label{app:extraction}

\begin{table}[H]
\centering
\caption{FARS (concept-centroid PCA, 10 dims) amplifies concept signal; control (form-centroid PCA, 5 dims) captures form. Boost = change in probe accuracy vs.\ full space.}
\label{tab:subspace_detail}
\small
\begin{tabular}{@{}lcccc|cc@{}}
\toprule
 & \multicolumn{4}{c|}{\textbf{FARS Subspace}} & \multicolumn{2}{c}{\textbf{Form Control}} \\
\textbf{Model} & RSA-C & RSA-F & Probe\% & Boost & RSA-F & RSA-C \\
\midrule
GPT-2 XL & .317 & .118 & 52.8 & +20.1 & .626 & $-$.003 \\
Qwen2.5-3B & .357 & .056 & 64.4 & +11.6 & .634 & $-$.005 \\
Qwen2.5-7B & .367 & .009 & 68.0 & +8.2 & .638 & $-$.010 \\
Mistral-7B & .366 & .008 & 70.5 & +10.2 & .631 & $-$.003 \\
Llama-3.1-8B & .368 & .015 & 71.9 & +18.0 & .634 & $-$.007 \\
\bottomrule
\end{tabular}
\end{table}

\section{Dimensionality Sweep Results}
\label{app:dimsweep}

\begin{table}[H]
\centering
\caption{Concept RSA at best layer as a function of subspace dimension $k$.}
\small
\begin{tabular}{@{}lccccc@{}}
\toprule
$k$ & \textbf{GPT-2 XL} & \textbf{Qwen-3B} & \textbf{Qwen-7B} & \textbf{Mistral} & \textbf{Llama-8B} \\
\midrule
1 & .196 & .250 & .273 & .271 & .265 \\
3 & .261 & .318 & .337 & .332 & .330 \\
5 & .291 & .341 & .356 & .351 & .354 \\
8 & .310 & .354 & .365 & .364 & .366 \\
10 & .317 & .357 & .367 & .366 & .368 \\
14 & .321 & .361 & .371 & .369 & .372 \\
17 & .323 & .363 & .373 & .371 & .374 \\
\bottomrule
\end{tabular}
\end{table}

\section{Cross-Model Alignment Details}
\label{app:alignment}

\begin{table}[H]
\centering
\caption{Pairwise FARS alignment for all 10 model pairs.}
\small
\begin{tabular}{@{}lcc@{}}
\toprule
\textbf{Model Pair} & \textbf{CCA mean} & \textbf{Centroid RSA} \\
\midrule
\multicolumn{3}{@{}l}{\emph{Same family}} \\
Qwen-3B vs.\ Qwen-7B & 0.92 & 0.93 \\
\midrule
\multicolumn{3}{@{}l}{\emph{Cross family}} \\
GPT-2 XL vs.\ Qwen-3B & 0.81 & 0.79 \\
GPT-2 XL vs.\ Qwen-7B & 0.83 & 0.82 \\
GPT-2 XL vs.\ Mistral-7B & 0.80 & 0.78 \\
GPT-2 XL vs.\ Llama-8B & 0.79 & 0.77 \\
Qwen-3B vs.\ Mistral-7B & 0.85 & 0.84 \\
Qwen-3B vs.\ Llama-8B & 0.84 & 0.82 \\
Qwen-7B vs.\ Mistral-7B & 0.88 & 0.87 \\
Qwen-7B vs.\ Llama-8B & 0.86 & 0.85 \\
Mistral-7B vs.\ Llama-8B & 0.90 & 0.89 \\
\bottomrule
\end{tabular}
\end{table}

\section{Leave-$K$-Out Generalization Results}
\label{app:generalization}

\begin{table}[H]
\centering
\caption{Leave-$K$-out generalization (10 random splits per $K$).}
\small
\begin{tabular}{@{}llcc@{}}
\toprule
\textbf{Model} & $K$ & \textbf{Held-out RSA} & \textbf{Held-out Probe\%} \\
\midrule
\multirow{3}{*}{GPT-2 XL} & 3 & $.39 \pm .04$ & $38.2 \pm 4.1$ \\
 & 6 & $.35 \pm .05$ & $33.7 \pm 5.2$ \\
 & 9 & $.28 \pm .06$ & $29.1 \pm 6.0$ \\
\midrule
\multirow{3}{*}{Mistral-7B} & 3 & $.42 \pm .03$ & $45.1 \pm 3.8$ \\
 & 6 & $.38 \pm .04$ & $40.3 \pm 4.5$ \\
 & 9 & $.33 \pm .05$ & $35.6 \pm 5.7$ \\
\midrule
\multirow{3}{*}{Llama-3.1-8B} & 3 & $.41 \pm .03$ & $43.8 \pm 3.5$ \\
 & 6 & $.37 \pm .04$ & $38.9 \pm 4.8$ \\
 & 9 & $.31 \pm .06$ & $33.2 \pm 5.4$ \\
\bottomrule
\end{tabular}
\end{table}

\section{Per-Domain Patching Analysis}
\label{app:perdomain}

\begin{table}[H]
\centering
\caption{FARS subspace patching overlap by reasoning domain (Mistral-7B, best layer, EN$\to$Math).}
\small
\begin{tabular}{@{}lcc@{}}
\toprule
\textbf{Domain} & \textbf{FARS Overlap} & \textbf{Full Overlap} \\
\midrule
Arithmetic & .968 & .583 \\
Logic & .961 & .571 \\
Relational & .952 & .548 \\
Causal & .947 & .562 \\
Spatial & .924 & .521 \\
\midrule
\textbf{Mean} & \textbf{.955} & \textbf{.557} \\
\bottomrule
\end{tabular}
\end{table}

\section{Supplementary Figures}
\label{app:supplementary_figures}

\begin{figure}[H]
\centering
\includegraphics[width=\linewidth]{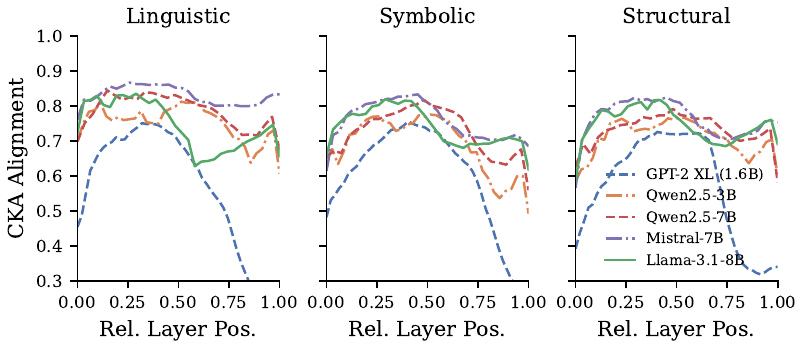}
\caption{CKA alignment by invariance dimension across layers. Linguistic, symbolic, and structural dimensions converge in middle layers.}
\label{fig:cka_dimensions}
\end{figure}

\begin{figure}[H]
\centering
\includegraphics[width=\linewidth]{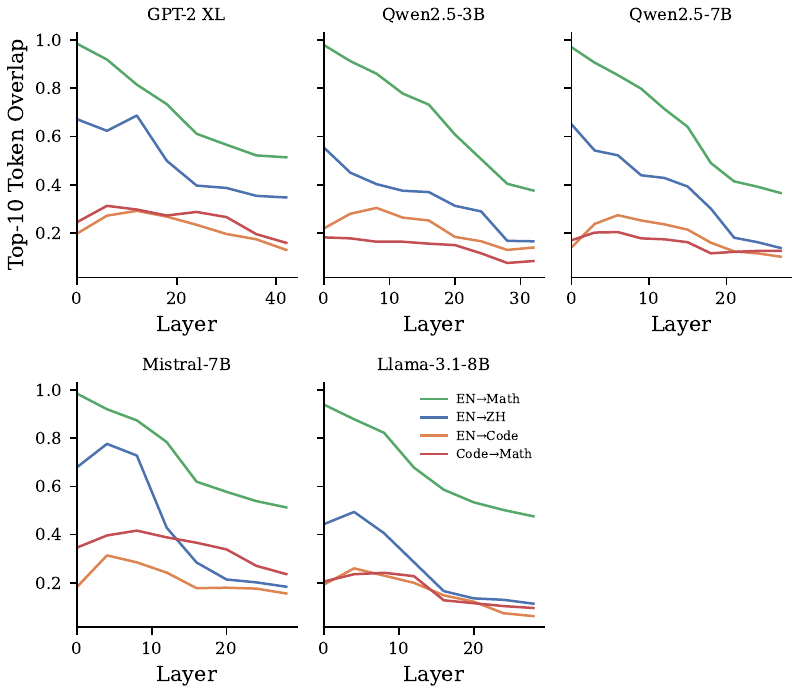}
\caption{Patching token overlap by layer, broken down by form pair. The declarative-procedural asymmetry is consistent across layers and models.}
\label{fig:patching_formpairs}
\end{figure}

\begin{figure}[H]
\centering
\includegraphics[width=\linewidth]{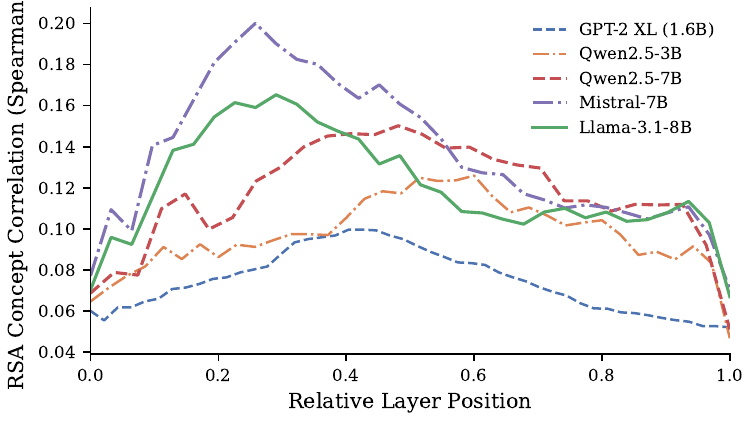}
\caption{Concept RSA across layers for all five models.}
\label{fig:rsa_concept_layers}
\end{figure}

\begin{figure}[H]
\centering
\includegraphics[width=\linewidth]{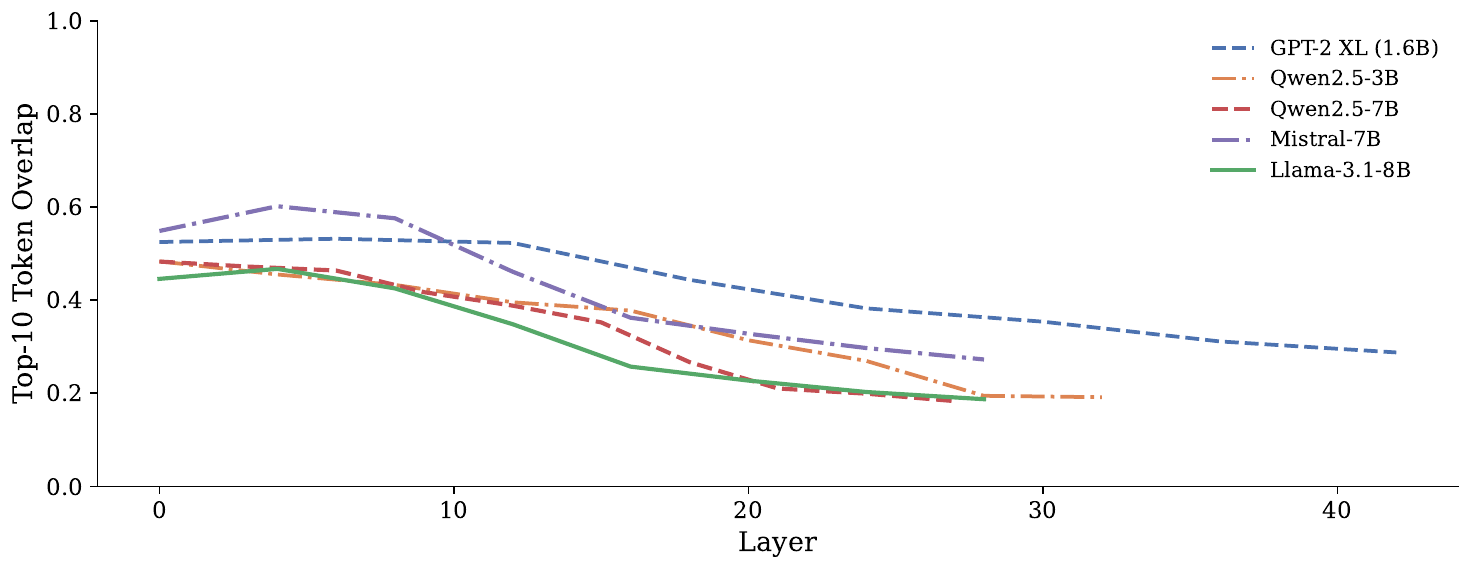}
\caption{Cross-form patching overlap by layer, averaged across all form pairs.}
\label{fig:patching_layers_app}
\end{figure}

\section{Implementation Details and Reproducibility}
\label{app:code}

\subsection{Environment and Models}

\begin{table}[H]
\centering
\caption{Software and hardware environment.}
\small
\begin{tabular}{@{}ll@{}}
\toprule
\textbf{Component} & \textbf{Version / Specification} \\
\midrule
Python & 3.10.12 \\
PyTorch & 2.1.0 (CUDA 12.1) \\
Transformers & 4.36.2 \\
scikit-learn & 1.3.2 \\
SciPy & 1.11.4 \\
\midrule
GPU & NVIDIA RTX 6000 Ada (48GB VRAM) \\
Inference precision & float16 \\
Analysis precision & float32 (CPU) \\
\bottomrule
\end{tabular}
\end{table}

All models are loaded from HuggingFace Hub in base (non-instruction-tuned) form: \texttt{openai-community/gpt2-xl}, \texttt{Qwen/Qwen2.5-3B}, \texttt{Qwen/Qwen2.5-7B}, \texttt{mistralai/Mistral-7B-v0.3}, \texttt{meta-llama/Llama-3.1-8B}.

\subsection{Hyperparameters}

\begin{table}[H]
\centering
\caption{Hyperparameter choices.}
\small
\begin{tabular}{@{}llp{5cm}@{}}
\toprule
\textbf{Parameter} & \textbf{Value} & \textbf{Justification} \\
\midrule
RSA permutations & 1{,}000 & Standard; $p < 0.001$ achievable \\
FDR correction & BH, $\alpha{=}0.05$ & Across all layers \\
Ridge probe $\alpha$ & 0.1 & Cross-validated \\
Bootstrap resamples & 5{,}000 & Concept-block resampling \\
FARS dimension $k$ & 10 & Elbow in dim.\ sweep (Fig.~\ref{fig:dim_sweep}) \\
Random baseline draws & 10 & QR-orthonormalized \\
Patching form pairs & 4 & EN$\to$Math, EN$\to$ZH, EN$\to$Code, Code$\to$Math \\
\bottomrule
\end{tabular}
\end{table}

\subsection{Compute Budget}

\begin{table}[H]
\centering
\caption{Compute cost breakdown.}
\small
\begin{tabular}{@{}lcc@{}}
\toprule
\textbf{Experiment} & \textbf{Per model} & \textbf{Total (5 models)} \\
\midrule
Activation extraction & ${\sim}$4 GPU-hrs & 20 GPU-hrs \\
Subspace patching (FARS) & ${\sim}$2 GPU-hrs & 10 GPU-hrs \\
Baseline patching (random + PCA) & ${\sim}$2 GPU-hrs & 10 GPU-hrs \\
CPU analyses & $<$12 min & $<$1 CPU-hr \\
\midrule
\textbf{Total} & & \textbf{${\sim}$40 GPU-hrs} \\
\bottomrule
\end{tabular}
\end{table}

\subsection{Core Implementation}

\noindent\textit{Activation extraction via forward hooks:}
\begin{lstlisting}[style=purplebox]
def extract_activations(model, tok, texts):
  all_acts = []
  for text in texts:
    inp = tok(text, return_tensors="pt").to(dev)
    captured = {}; hooks = []
    for i, ly in enumerate(model.model.layers):
      def hook(mod, _in, out, idx=i):
        captured[idx] = out[0][0,-1].cpu()
      hooks.append(ly.register_forward_hook(hook))
    with torch.no_grad(): model(**inp)
    for h in hooks: h.remove()
    all_acts.append(torch.stack(
      [captured[i] for i in range(len(captured))]))
  return torch.stack(all_acts).numpy()
\end{lstlisting}

\noindent\textit{FARS extraction via concept-centroid PCA:}
\begin{lstlisting}[style=purplebox]
def extract_fars(activations, c_labels, k=10):
  N, L, D = activations.shape
  basis = np.zeros((L, k, D))
  for layer in range(L):
    X = activations[:, layer, :]
    centroids = np.stack([
      X[c_labels == c].mean(0)
      for c in np.unique(c_labels)])
    pca = PCA(n_components=k).fit(centroids)
    basis[layer] = pca.components_
  return basis
\end{lstlisting}

\noindent\textit{Subspace patching} (Eq.~\ref{eq:subspace_patch}):
\begin{lstlisting}[style=purplebox]
def subspace_patch(h_src, h_tgt, B):
  # B: (k, D) orthonormal FARS basis
  diff = h_src - h_tgt
  return h_tgt + B.T @ (B @ diff)
\end{lstlisting}

\noindent\textit{Subspace ablation:}
\begin{lstlisting}[style=purplebox]
def subspace_ablate(h, B):
  return h - B.T @ (B @ h)
\end{lstlisting}

\noindent\textit{Permutation-based RSA:}
\begin{lstlisting}[style=purplebox]
def perm_rsa(emp_rdm, theo_rdm, n=1000):
  N = emp_rdm.shape[0]
  ix = np.triu_indices(N, k=1)
  obs = spearmanr(emp_rdm[ix], theo_rdm[ix])[0]
  count = 0
  for _ in range(n):
    p = np.random.permutation(N)
    pr = theo_rdm[np.ix_(p, p)]
    if spearmanr(emp_rdm[ix], pr[ix])[0] >= obs:
      count += 1
  return obs, (count + 1) / (n + 1)
\end{lstlisting}

\noindent\textit{Concept-block bootstrap for confidence intervals:}
\begin{lstlisting}[style=purplebox]
def bootstrap_ci(values, groups, n=5000):
  unique = np.unique(groups)
  boots = np.zeros(n)
  for b in range(n):
    idx = np.random.choice(unique,
      size=len(unique), replace=True)
    vals = np.concatenate(
      [values[groups == g] for g in idx])
    boots[b] = np.mean(vals)
  return np.percentile(boots, [2.5, 97.5])
\end{lstlisting}

\end{document}